\documentclass{article}
\usepackage{iclr2025_conference,times}

\usepackage[utf8]{inputenc} % allow utf-8 input
\usepackage[T1]{fontenc}    % use 8-bit T1 fonts
\usepackage{hyperref}       % hyperlinks
\usepackage{url}            % simple URL typesetting
\usepackage{booktabs}       % professional-quality tables
\usepackage{amsfonts}       % blackboard math symbols
\usepackage{nicefrac}       % compact symbols for 1/2, etc.
\usepackage[verbose]{microtype}      % microtypography
\usepackage{xcolor}         % colors

\usepackage{kotex}
\usepackage[linesnumbered,ruled]{algorithm2e}
\usepackage{algpseudocode}
\usepackage{algcompatible}
\usepackage[nointegrals]{wasysym}
\usepackage{amsmath}
\usepackage{float}
\usepackage[pdftex]{graphicx}
\usepackage{caption}
\usepackage{subcaption}
\usepackage{amsmath}
\usepackage{amssymb}
\usepackage{bm}
\newcommand{\argmax}{\mathop{\mathrm{argmax}}}

\title{Diffusion-Based Offline RL for Improved Decision-Making in Augmented ARC Task
}

\author{Yunho Kim\textsuperscript{1} \enspace Jaehyun Park\textsuperscript{1} \enspace Heejun Kim\textsuperscript{1} \enspace Sejin Kim\textsuperscript{1} \enspace Byung-Jun Lee\textsuperscript{2} \enspace Sundong Kim\textsuperscript{1}
\\  \textsuperscript{1}Gwangju Institute of Science and Technology \qquad \textsuperscript{2}Korea University \\
}

\begin{document}

%if final version
\iclrfinalcopy

% Below three lines are for ArXiv - to remove the header 
\pagestyle{plain}
\thispagestyle{plain}
\renewcommand{\headrulewidth}{0pt}

\maketitle

\begin{abstract}
\label{Sec:Abstract}

Effective long-term strategies enable AI systems to navigate complex environments by making sequential decisions over extended horizons. Similarly, reinforcement learning (RL) agents optimize decisions across sequences to maximize rewards, even without immediate feedback. To verify that Latent Diffusion-Constrained Q-learning (LDCQ), a prominent diffusion-based offline RL method, demonstrates strong reasoning abilities in multi-step decision-making, we aimed to evaluate its performance on the Abstraction and Reasoning Corpus (ARC). However, applying offline RL methodologies to enhance strategic reasoning in AI for solving tasks in ARC is challenging due to the lack of sufficient experience data in the ARC training set. To address this limitation, we introduce an augmented offline RL dataset for ARC, called Synthesized Offline Learning Data for Abstraction and Reasoning (SOLAR), along with the SOLAR-Generator, which generates diverse trajectory data based on predefined rules. SOLAR enables the application of offline RL methods by offering sufficient experience data. We synthesized SOLAR for a simple task and used it to train an agent with the LDCQ method. Our experiments demonstrate the effectiveness of the offline RL approach on a simple ARC task, showing the agent's ability to make multi-step sequential decisions and correctly identify answer states. These results highlight the potential of the offline RL approach to enhance AI's strategic reasoning capabilities.

\end{abstract}

\section{Introduction}
\label{Sec:Intro}
% shortened introduction, to fit up to section 2.1 into the second page.

% A deliberate thinking process,often referred to as System-2 reasoning, involves careful, deliberate thought in evaluating options and determining the best course of action. This type of reasoning is require requires conscious effort, allowing intelligent beings to systematically plan and execute multi-step strategies to achieve complex, long-term objectives. Reinforcement learning (RL) shares similarities with this process, as RL agents make decisions over extended sequences to maximize rewards without immediate feedback. In both cases, reasoning involves considering the extended actions in order to reach an optimal outcome. The way the Q-value guides the agent toward desired outcomes aligns with System-2 reasoning's requirement to make multi-step decisions to achieve a goal. Enhancing AI's ability to make a decision for the future can thus improve its performance in navigating complex environments and achieving long-term objectives, aligning RL with System-2 reasoning.

Effective long-term strategies involve deliberate reasoning, which refers to the thoughtful evaluation of options to determine the best course of action~\citep{kahneman2011think}. This type of reasoning requires conscious effort and allows intelligent beings to systematically plan and execute multi-step strategies to achieve complex long-term goals. Similarly, reinforcement learning (RL) agents make decisions with the goal of maximizing rewards over extended sequences of actions, even without immediate feedback. In both cases, reasoning involves considering a sequence of actions to reach an optimal outcome. We believe that the way Q-values guide an RL agent toward desired outcomes aligns with the subgoals of deliberate reasoning, particularly in terms of multi-step decision-making to achieve long-term objectives.

Recent approaches to offline RL combined with generative diffusion models have shown significant improvements in multi-step strategic decision-making abilities for future outcomes~\citep{janner2022planning,ajay2022conditional,liang2023adaptdiffuser,li2023hierarchical}. In particular, Latent Diffusion-Constrained Q-learning (LDCQ)~\citep{venkatraman2024reasoning} leverages diffusion models to sample various latents that compress multi-step trajectories. These latents are then used to guide the Q-learning process. By generating diverse data based on in-distribution samples, diffusion models help overcome the limitations of fixed datasets. This integration of diffusion models into offline RL enhances agents' reasoning abilities, allowing them to consider multiple plausible trajectories across extended sequences.

\begin{figure}[h]
  \centering
  \includegraphics[width=.72\linewidth]{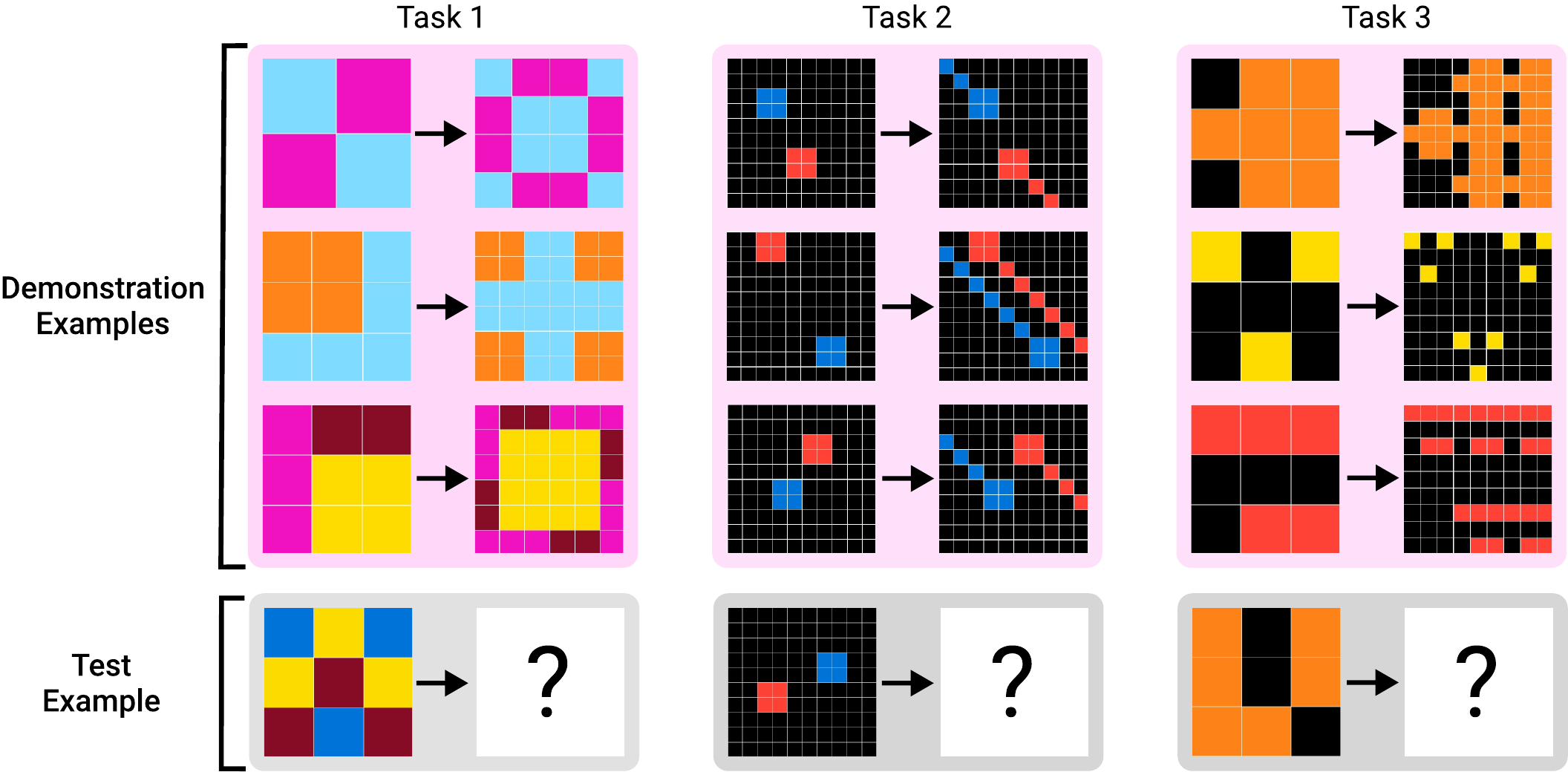}
  \caption{\normalsize Three tasks in ARC. Each task consists of demonstration examples and a test example. Each example has an input grid and an output answer grid. Each pixel in the grid is matched to a color corresponding to a value in the range 0–9. ARC requires identifying common rules from the demonstration examples and applying them to solve the test example correctly. Despite recent advancements in AI, current models have consistently underperformed compared to humans on the ARC benchmark~\citep{arcprize2024,johnson2021fast}.}
  \label{ARC-Example}
\end{figure}

To rigorously evaluate whether offline RL methods possess the advanced reasoning abilities required to solve complex tasks, we chose the Abstraction and Reasoning Corpus (ARC)~\citep{chollet2019ARC}, one of the key benchmarks for measuring the abstract reasoning ability in AI. As shown in Figure~\ref{ARC-Example}, the ARC training set consists of 400 grid-based tasks, each requiring the identification of common rules from demonstration examples, which are then applied to solve the test examples. ARC tasks are particularly challenging for AI models because they demand high-level reasoning abilities, integrating core knowledge priors such as objectness, basic geometry, and topology~\citep{chollet2019ARC}. These core knowledge priors guide the decision-making process for selecting the appropriate actions. Therefore, we believe that agents trained with offline RL methods can leverage these core knowledge priors by learning from the experienced data. 

However, the existing ARC training dataset lacks sufficient trajectories for training agents with offline RL methods. To address this limitation, we propose Synthesized Offline Learning data for Abstraction and Reasoning (SOLAR), a dataset for training offline RL agents. SOLAR provides diverse trajectory data, allowing the agent to encounter various actions shaped by the core knowledge priors across different episodes. In this research, we synthesized SOLAR for a simple task using the SOLAR-Generator, which creates data according to the desired conditions. The synthesized SOLAR was then used to train agents using the LDCQ method. Through training with LDCQ on SOLAR, agents demonstrated the ability to devise pathways to correct answer states, even generating solution paths not present in the training data. This approach highlights the potential of diffusion-based offline RL methods to enhance AI’s reasoning capabilities.

% Despite recent advancements in AI, current models have consistently underperformed compared to humans on the ARC benchmark~\citep{kaggle2024arc,johnson2021fast}. This performance gap highlights the need for advanced approaches to enhance AI’s reasoning abilities, particularly in abstract and multi-step decision-making. We believe that diffusion-based offline RL methods can help to reduce this gap by training the experienced trajectories that they show stitching ability in long-horizon sparse reward environments. However, the existing ARC dataset does not provide enough diverse trajectories that agents can learn from to effectively apply offline RL methods.

% This can be achieved by training a $\beta$-Variational autoencoder (VAE)~\citep{higgins2017beta,kingma2013auto}, which captures how previous segmented trajectories work by encoding them into latent representations. 

\section{Preliminaries}

\subsection{ARC Learning Environment (ARCLE)}
\label{Sec:ARCLE}
% \sd{ARCLE 논문 참고하여 paraphrasing해도 됩니다: \url{https://www.overleaf.com/read/grvrdftkhzqk#d67ef6}}

\begin{figure}[htb!]
  \centering
  \includegraphics[width=1\linewidth]{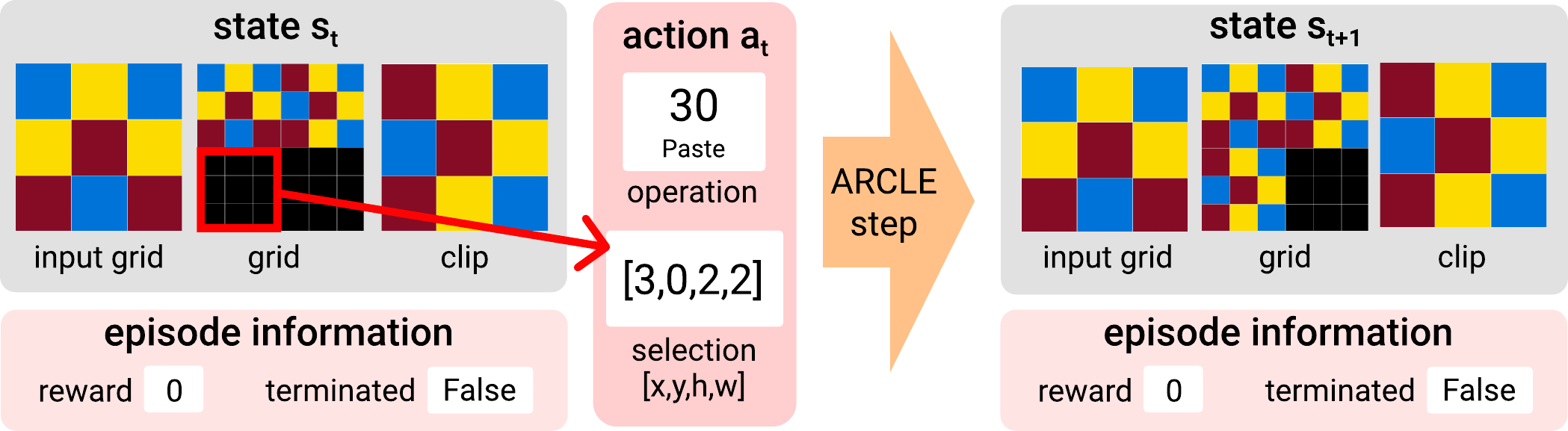}
  \caption{\normalsize An example of a single step in ARCLE. In this example step, the action has an operation 30 (\texttt{Paste}) and a selection of $[3, 0, 2, 2]$. The top-left coordinate of the selection box is $[3,0]$ and the bottom-right coordinate is $[5,2]$. $[h_t,w_t]$ is calculated by subtracting $[3,0]$ from $[5,2]$. When ARCLE executes this action, the current clipboard is pasted into the bounding box specified by the selection on the current grid. It then returns episode information, including the reward and termination status.}
  \label{simple_example_ARCLE}
\end{figure}

ARCLE~\citep{lee2024arcle} is a Gymnasium-based environment developed to facilitate RL approaches for solving ARC tasks. ARCLE frames ARC tasks within a Markov Decision Process (MDP) structure, providing an environment where agents can interact with and manipulate grid-based tasks. This MDP structure enables ARC tasks to be solved through sequential decision-making. 

ARCLE handles states and actions following the O2ARC web interface~\citep{shim2024o2arc}. As shown in Figure~\ref{simple_example_ARCLE}, when ARCLE executes an action $\bm{a}_t$ on the current state $\bm{s}_t$, it returns the next state $\bm{s}_{t+1}$, along with episode information about the reward and termination status. A state $\bm{s}_t$ consists of $(\text{input grid}, \text{grid}_t, \text{clipboard}_t)$ at timestep $t$. The input grid represents the initial state of the test example, the $\text{grid}_t$ denotes the current grid at time $t$ after several actions have been applied, and the $\text{clipboard}_t$ stores the copied grid by the \texttt{Copy} operation. An action $\bm{a}_t$ consists of $(\text{operation}_t, x_t, y_t, h_t, w_t)$, where $\text{operation}_t$ represents the type of transformation, $x_t$ and $y_t$ denote the coordinates of the top-left point of the selection box, and $h_t$ and $w_t$ represent the difference between the bottom-right and top-left coordinates. All subsequent notations for $\bm{s}_t$ and $\bm{a}_t$ will adhere to this definition for clarity. Reward is only given when the \texttt{Submit} operation is executed at the answer state, and the episode terminates either after receiving the reward or when \texttt{Submit} is executed across multiple trials. All possible operations are mentioned in Appendix~\ref{Appendix:operations in SOLAR}. 

% where agents interact with the environment by selecting actions that manipulate grid-based tasks. By providing a structured environment for grid-based tasks, ARCLE simplifies the process of creating RL models that can handle diverse problem sets, allowing for more efficient testing and development of algorithms aimed at solving ARC tasks.

% ARCLE effectively tackles the challenges posed by ARC’s vast action space and diverse task types by providing an environment where RL agents can develop robust planning strategies. Through skill abstraction and structured actions, ARCLE allows agents to break down complex problems into smaller, manageable steps, making it easier to navigate the large solution space. This structured approach enhances the agent’s ability to generalize across tasks, making ARCLE a valuable tool for advancing reasoning and planning capabilities in reinforcement learning.

\subsection{Diffusion-Based Offline Reinforcement Learning}

Offline RL focuses on learning policies from previously collected data, without interacting with the environment. However, Offline RL faces challenges, including data distribution shifts, limited diversity in the collected data, and the risk of overfitting to biased or insufficiently representative samples. To address these issues, several works in offline RL have focused on improving learning efficiency with large datasets and enhancing generalization to unseen scenarios while balancing diversity and ensuring data quality~\citep{fujimoto2019off, kidambi2020morel, levine2020offline}.

Recent offline RL methods offer promising solutions in long-horizon tasks and handling out-of-support samples through diffusion models. For instance, Diffuser~\citep{janner2022planning} generates tailored trajectories by learning trajectory distributions, reducing compounding errors. Beyond this, a range of advanced diffusion-based offline RL approaches, such as Decision Diffuser (DD)~\citep{ajay2022conditional}, AdaptDiffuser~\citep{liang2023adaptdiffuser}, HDMI~\citep{li2023hierarchical}, have demonstrated the effectiveness of combining diffusion models with offline RL.

% Therefore, we selected the LDCQ method to evaluate the reasoning ability of offline RL agents in solving ARC tasks.
\subsection{Latent Diffusion-Constrained Q-Learning (LDCQ)}

\begin{figure}[ht]
    \centering
    \begin{subfigure}{0.40\textwidth}
        \centering
        \includegraphics[width=\linewidth]{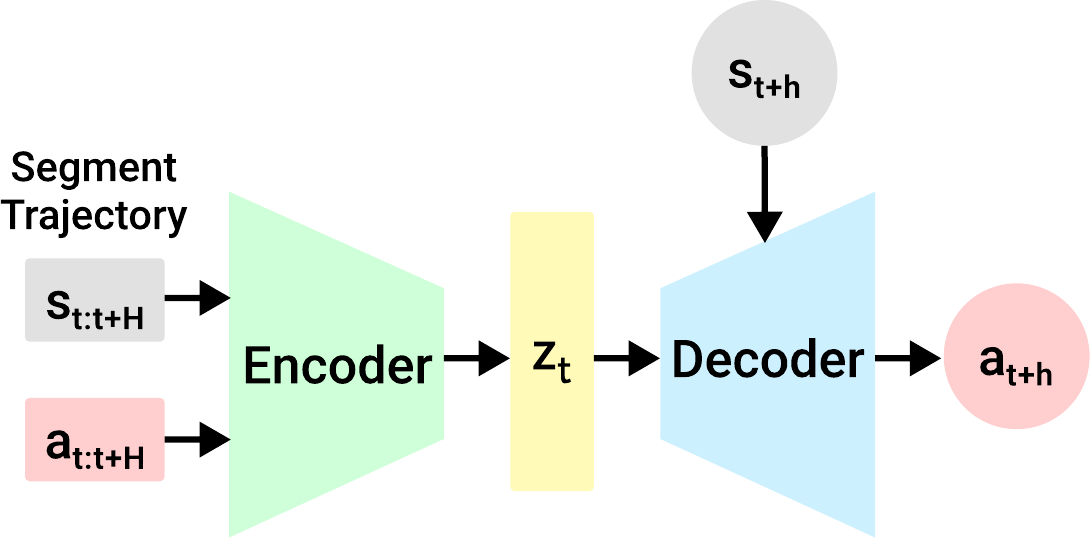}
        \caption{\normalsize Training $\beta$-VAE}
        \label{fig:ldcq-a}
    \end{subfigure}
    \hspace{0.04\textwidth}
    \begin{subfigure}{0.40\textwidth}
        \centering
        \includegraphics[width=\linewidth]{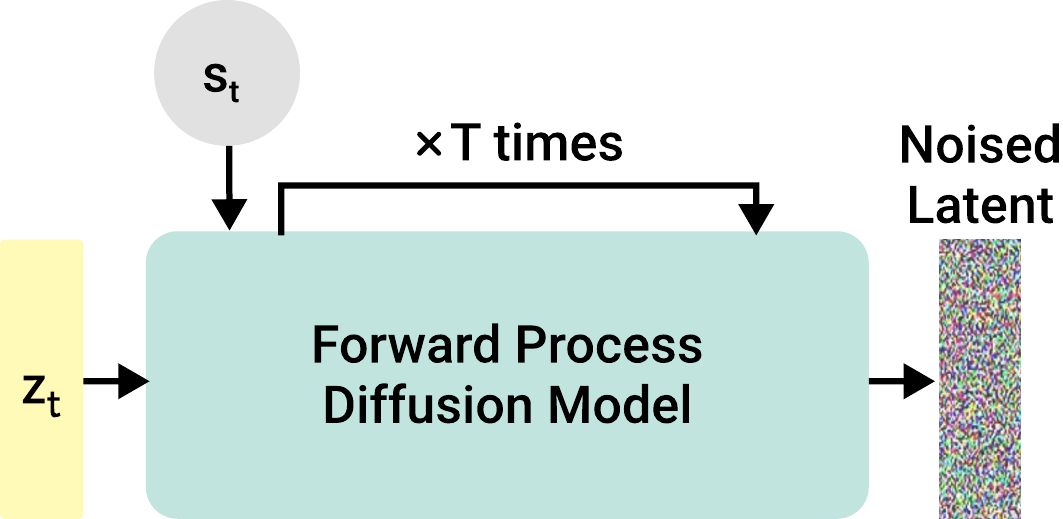}
        \caption{\normalsize Training Latent Diffusion Model}
        \label{fig:ldcq-b}
    \end{subfigure}
    \hspace{0.04\textwidth}
    \begin{subfigure}{0.40\textwidth}
        \centering
        \includegraphics[width=\linewidth]{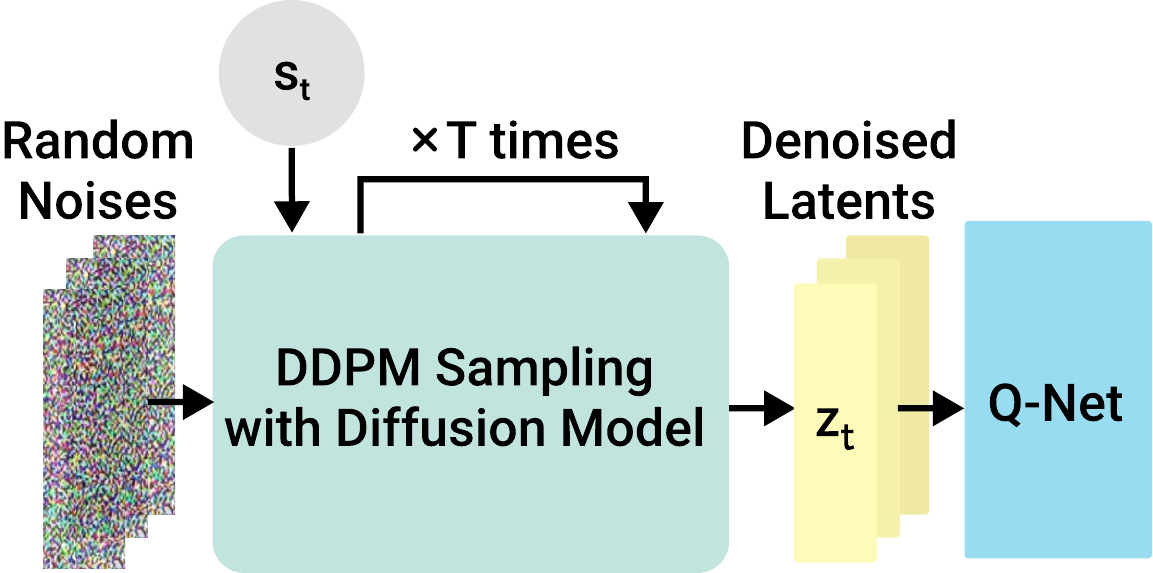}
        \caption{\normalsize Training Q-Network}
        \label{fig:ldcq-c}
    \end{subfigure}
    \hspace{0.04\textwidth}
    \begin{subfigure}{0.40\textwidth}
        \centering
        \includegraphics[width=\linewidth]{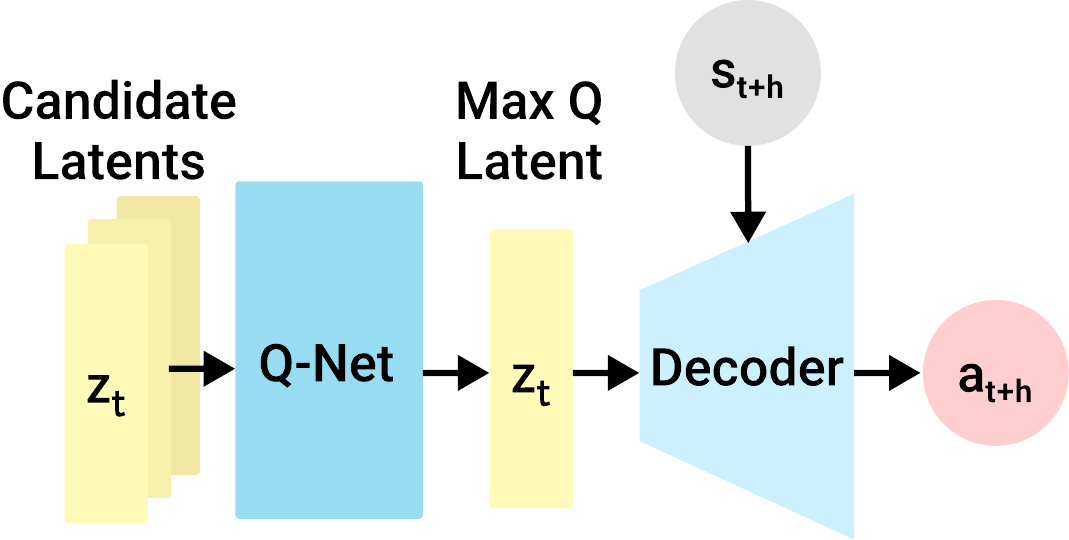}
        \caption{\normalsize Inference step}
        \label{fig:ldcq-d}
    \end{subfigure}
    
    \caption{\normalsize (a)--(c) Training stages of LDCQ. (a) Training a $\beta$-VAE with an encoder that encodes $\mathit{H}$-horizon segment trajectories into latents $\bm{z}_t$, and a policy decoder that decodes actions based on $\bm{z}_t$ and state $\bm{s}_{t+h}$ where $h \in [ 0,H )$ contained in the latent. (b) Training a diffusion model based on $\bm{z}_t$ and the $\bm{s}_t$. (c) Training a Q-network using latents sampled through the diffusion model. (d) LDCQ inference step at $\bm{s}_{t+h}$. Possible latents at $\bm{s}_t$ are sampled through the diffusion model, and the agent executes actions resulting from decoding the latent with the highest Q-value.}
    \label{fig:LDCQ}
\end{figure}

Latent Diffusion-Constrained Q-learning (LDCQ)~\citep{venkatraman2024reasoning} leverages latent diffusion and batch-constrained Q-learning to handle long-horizon, sparse reward tasks more effectively. The LDCQ method uses sampled latents that encode trajectories of length $H$ to train the Q-function, effectively reducing extrapolation error. The training process of LDCQ is shown in Figure~\ref{fig:LDCQ}: 1) training the $\beta$-VAE to learn latent representations, 2) training the diffusion model using the latent vectors encoded by the $\beta$-VAE, and 3) training the Q-network with latents sampled from the diffusion model. More details about the LDCQ method are described in Appendix~\ref{Appendix:Training Details}.

\section{Synthesized Offline Learning data for Abstraction and Reasoning (SOLAR)}
% 어떤 내용이 챕터 3에 담기게 될 지 1-2문장 정도로 요약하거나, 제안하는 방법론에 도달하게 된 insight를 1문단 정도로 남길 수 있는 공간

We developed a new dataset called Synthesized Offline Learning data for Abstraction and Reasoning (SOLAR) that can be used to train offline RL methods. Solving ARC tasks can be considered a process of making multi-step decisions to transform the input grid into the output answer grid. We believe that the process of making these decisions inherently involves applying core knowledge priors, objectness, goal-directedness, numbers and counting, and basic geometry and topology~\citep{chollet2019ARC}, which are necessary for solving ARC tasks. The ARC training set lacks information on how to solve the task, and it only provides a set of demonstration examples and a test example for each task, as shown in Figure~\ref{ARC-Example}. To address this, We aim to provide the trajectory data to solve the task through SOLAR, enabling them to learn how actions change the state based on the application of core knowledge priors.

\subsection{SOLAR Structure}
SOLAR contains various transition data $\left(\bm{s}_t, \bm{a}_t, \bm{s}_{t+1}\right)$, where actions $\bm{a}_t$ are taken in different states $\bm{s}_t$, and the result $\bm{s}_{t+1}$ observed. To facilitate effective learning and a combination of core knowledge, we use ARCLE~\citep{lee2024arcle}. By designing a simple reward system that only provides rewards upon reaching the correct solution, we can guide the agent towards the desired state using reinforcement learning methods. 

\begin{figure}[h]
  \centering
  \includegraphics[width=1.0\linewidth]{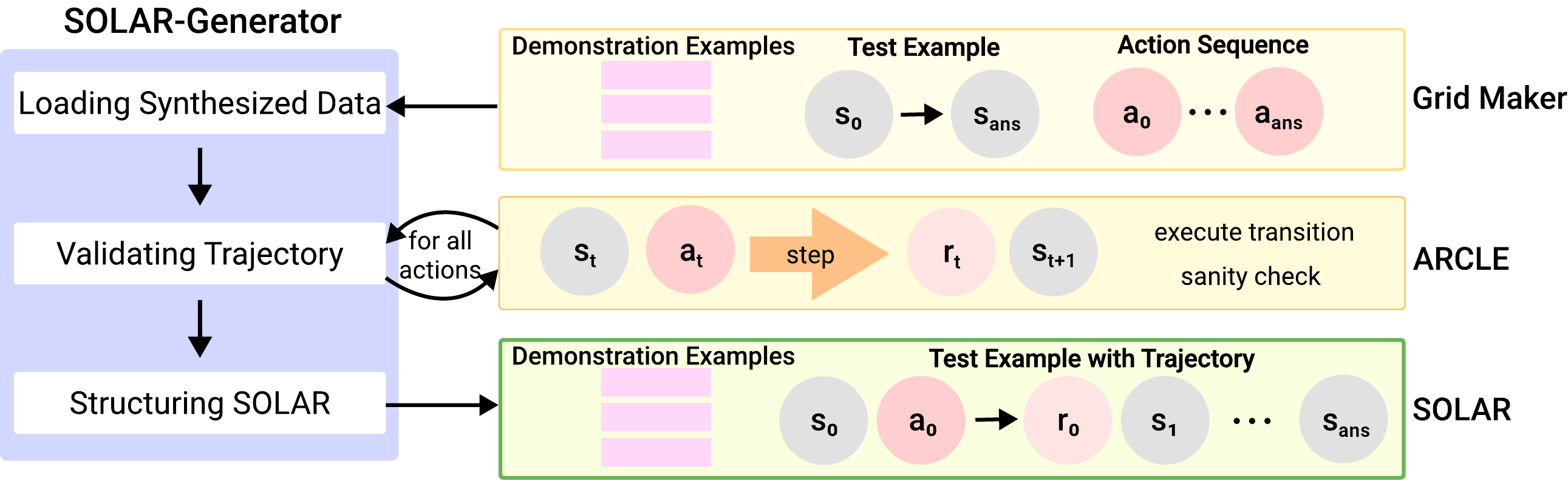}
  \caption{Data synthesis procedure with SOLAR-Generator. The state and actions consist of as mentioned in Section~\ref{Sec:ARCLE}. 1) Loading Synthesized Data: The Grid Maker module applies constraints, augments input-output pairs, and synthesizes solutions for specific tasks by utilizing actions. 2) Validating Trajectories: Checks whether the generated actions are executable in ARCLE. 3) Structuring SOLAR: Organizes and stores the synthesized data in SOLAR based on the defined format. This step determines what information to include in the dataset and whether to segment episodes into fixed-length chunks or store them as a whole.}
  \label{fig:SOLAR-Generator}
\end{figure}

As shown in Figure~\ref{fig:SOLAR-Generator}, SOLAR consists of two key components: \emph{Demonstration Examples} and \emph{Test Example with Trajectory}. The demonstration examples and the test examples serve the same roles as in ARC. Through the demonstration examples, the common rule for transforming the input grid to the output grid is identified and then applied to solve the test example. Trajectory data means the episode data that starts from test input $s_0$.

\subsection{SOLAR-Generator}

We developed the SOLAR-Generator to synthesize SOLAR. SOLAR-Generator augments ARC trajectories by following ARCLE formalism, addressing the inherent complexity and diversity of ARC tasks. Figure~\ref{fig:SOLAR-Generator} illustrates the data synthesis procedure, which is carried out in three steps: 1) Loading Synthesized Data, 2) Validating Trajectories with ARCLE, and 3) Structuring SOLAR.

\paragraph{Loading Synthesized Data} The first step in SOLAR-Generator is to load the synthesized data for the target tasks. SOLAR provides the Grid Maker with common parameters such as maximum grid size and the number of demonstration examples per test example. Each task has its own specific Grid Maker, which synthesizes demonstration examples, test examples, and corresponding action sequences (selections and operations) based on the task’s constraints and rules. If desired, non-optimal trajectories containing random actions can also be synthesized. At this stage, the Grid Maker synthesizes only grid pairs and possible action sequences. The full trajectory data for the test example is constructed after passing through ARCLE. More details about how the Grid Maker synthesizes the input-output grids and action sequences are described in Appendix~\ref{Appendix:SOLAR}.

\paragraph{Validating Trajectories with ARCLE} After synthesizing various grids and action sequences with the Grid Maker, SOLAR-Generator checks whether the action sequences are valid in ARCLE. The Grid Maker serves as a data loader, enabling it to load and validate the synthesized data. Through this process, ARCLE provides intermediate states, rewards, and termination status for each step, and verifies that each action is correctly executed in the current state. This step is particularly important for non-optimal trajectories, where operations and selections may be generated randomly, as invalid selections can sometimes be synthesized by the Grid Maker. For \emph{gold standard} trajectories, intended as correct solutions, SOLAR-Generator ensures that the final grid of the trajectory matches the expected output grid of the test example. As a result, this stage is useful for checking and debugging the synthesized trajectories, preventing unintended errors.

\paragraph{Structuring SOLAR} After the trajectory validation is complete, the episodes are saved into SOLAR. In this step, user can determine the necessary information to include in SOLAR. At its core, SOLAR includes episodes consisting of state, action, reward, and termination information at each step, which are essential for training with offline RL methods. In addition to the previously mentioned information, SOLAR can also store various data from ARCLE, such as grid sizes at each step, binary mask versions of selections, and other relevant information needed for different learning methods. In this research, we designed the data to work with methods like LDCQ, which require trajectories of fixed horizon length $\mathit{H}$. Therefore, the trajectories are segmented into fixed-length chunks with a horizon length of $\mathit{H}$.

Through these three steps, SOLAR-Generator synthesizes diverse solutions by altering action orders or using alternative operation combinations. This is achieved by the Grid Maker, which generates data using pre-implemented algorithms, enabling the user to create as many trajectories as needed. SOLAR provides a sufficient training set for learning various problem-solving strategies. By offering diverse trajectories while adhering to the task-solving criteria, SOLAR bridges the gap between ARC's reasoning challenges and the sequential decision-making process of offline RL. For additional details about SOLAR and SOLAR-Generator, see the project website\footnote{\url{https://github.com/GIST-DSLab/SOLAR-Generator}} and Appendix~\ref{Appendix:SOLAR}.

\section{Design SOLAR for a Simple Task}
\label{Sec:SOLAR_for_single_task}

One of the most crucial factors in solving ARC tasks is the ability to recognize whether the current state is the answer state and to submit the correct answer accordingly. In ARC, each task embodies a single analogy, but this analogy can be approached through various action sequences~\citep{johnson2021fast, kim2022playgrounds}. Some solution paths may better exemplify the underlying analogy, while others might be less optimal or clear \citep{kim2022playgrounds}. Moreover, even when solving different test examples within the same task where the same rule is applied, the actual action sequence can vary depending on factors like the grid size or the arrangement of elements in the input grid. The diversity in potential action sequences to solve a single ARC task highlights the complexity of abstract reasoning and the importance of identifying the core analogy. 
%Furthermore, the answer is determined based on core knowledge priors such as objectness, geometry, and topology~\cite{chollet2019ARC}. %The input grids are not the same, so the answer grids are also not the same even within the same task. 
Therefore, an agent's ability to judge that it has reached an answer state implies that it has comprehended the underlying analogy and executed the necessary ARCLE actions to arrive at the correct solution. This ability to recognize the answer state is critical, as it demonstrates the agent's understanding of the task's inherent logic and its capacity to apply appropriate problem-solving strategies. In ARCLE, the reward is given only when the agent predicts the \texttt{Submit} operation and the submitted grid is the same as the answer grid. To evaluate whether an agent trained with LDCQ can correctly identify and submit at the answer state—even when non-optimal trajectories are included in the training dataset—we mixed in incorrect episodes where the \texttt{Submit} operation is conducted in non-answer states.

Given these characteristics of ARC tasks, our experimental objectives are: 1) To assess whether the model can reach the answer state when various non-optimal trajectories are mixed with gold standard trajectories, and 2) To determine whether the model can recognize the answer state and perform the \texttt{Submit} action appropriately. By demonstrating the model's ability to identify the answer state, we can infer that it has internalized core knowledge priors and understands the high-level problem-solving methods necessary for ARC tasks. We synthesized SOLAR for a simple task designed to show these experimental objectives.

% \subsection{SOLAR for simple task}

\begin{figure}[h]
  \centering
  \includegraphics[width=1.0\linewidth]{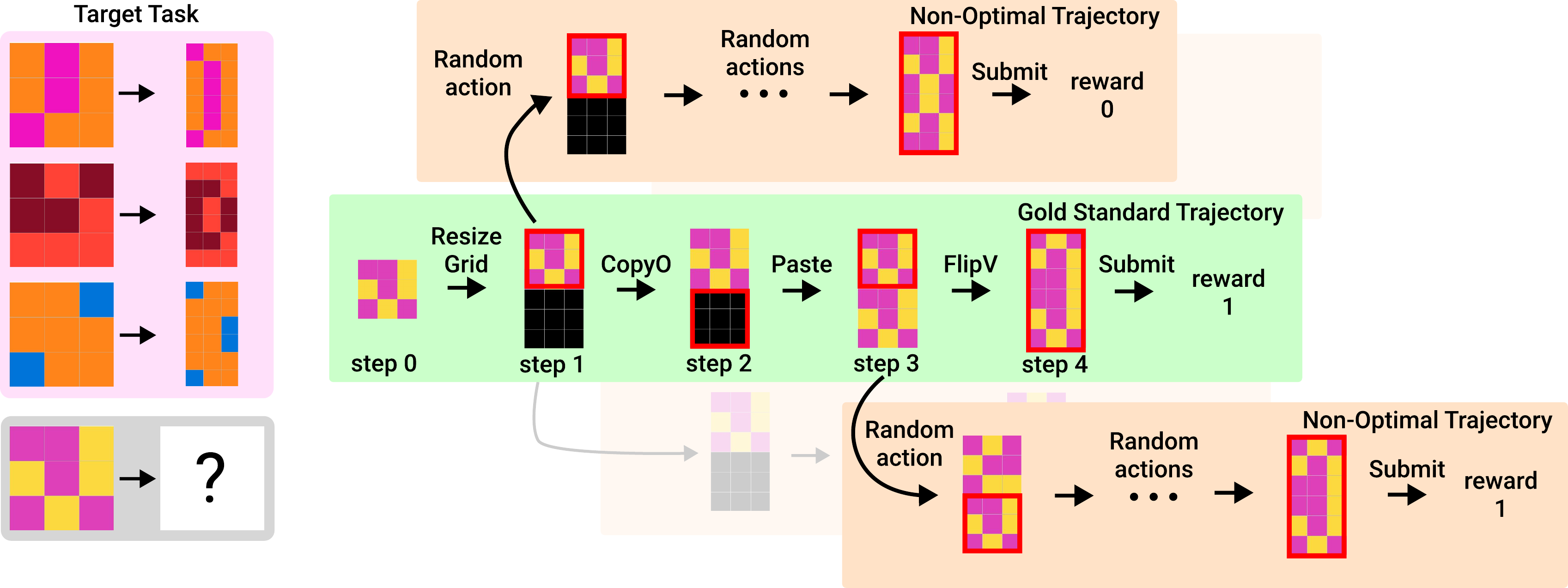}
  \caption{SOLAR episodes for a simple task: The gold standard trajectory (episode) contains the steps to solve the problem by using the core knowledge priors properly. The non-optimal episodes branch off at a random step within the standard trajectory, performing random operations such as \texttt{Rotate}, \texttt{Flip}, or \texttt{Copy} \& \texttt{Paste}, and then \texttt{Submit} after a certain number of steps.}
  \label{fig:task}
\end{figure}

We designed a simple task that requires core knowledge priors such as objectness and geometry. This task necessitates the ability to consider the input grid as an object and then perform actions based on this object. We constrained the maximum grid size to 10x10, and each episode includes three demonstration pairs. In creating SOLAR for this task, we constructed the dataset to include both gold standard episodes—which successfully reach the answer state and perform the \texttt{Submit} action—and non-optimal episodes—which follow random paths that may or may not reach the answer state. The inclusion of non-optimal trajectories was intended to evaluate whether the agent can recognize the answer state and appropriately perform the \texttt{Submit} action, thereby assessing its reasoning abilities rather than simply mimicking the actions in the dataset. As shown in Figure~\ref{fig:task}, the gold standard episode for this task consists of 5 steps: 1) \texttt{ResizeGrid} to make the grid two times longer vertically, 2) \texttt{CopyO} to copy the upper half of the current grid, as it matches the input grid, 3) \texttt{Paste} to apply it to the lower half of the grid, 4) \texttt{FlipV} to vertically flip the upper half of the current grid, and 5) \texttt{Submit}, as it reaches the answer state.

In the non-optimal episodes, the trajectories initially follow the gold standard trajectory but deviate at a random step to execute random actions for several steps. We constrained the random operations to \texttt{FlipV} (vertical flip), \texttt{FlipH} (horizontal flip), \texttt{Rotate90} (counterclockwise rotation), \texttt{Rotate270} (clockwise rotation), and \texttt{CopyO} (updating the clipboard with the selected area). For selection, it was constrained to either two options (upper half or lower half of the current grid) or three options (upper half, lower half, or the whole grid). Specifically, there are two options for \texttt{Rotate90}, \texttt{Rotate270}, and \texttt{CopyO}, and three options for the others. When \texttt{CopyO} is selected, the subsequent \texttt{Paste} action is forced onto the other possible selection option. This simplified selection allows for focusing on assessing the AI's decision-making by sequentially combining operations.

Each non-optimal episode contains approximately ten steps to the end, allowing the trajectory to include various actions in diverse states. For each problem pair, one gold standard episode and nine non-optimal episodes were generated, totaling 5,000 episodes across 500 problem pairs. As a result, the training set was composed such that approximately 10\% of the total episodes included the \texttt{Submit} operation at the answer state.

\section{Experiments and Results}
\label{Sec:Result}

\subsection{Evaluation Process using ARCLE}
\begin{figure}[ht]
  \centering
  \includegraphics[width=.9\linewidth]{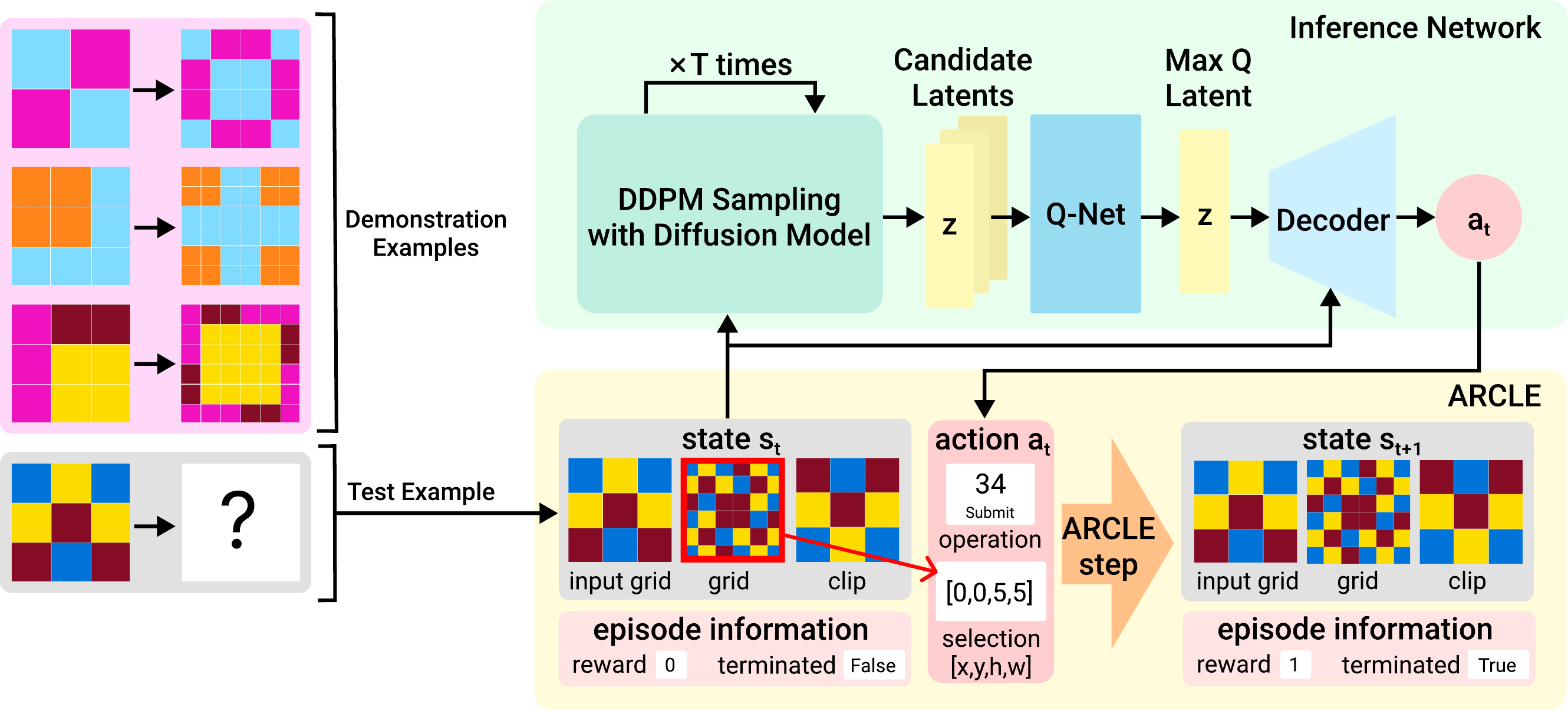}
  \caption{Inference framework for solving ARC tasks. ARCLE loads the task from the dataset and manages state information as well as the termination status of the current evaluation episode. The inference network of LDCQ performs DDPM sampling on the given state to extract candidate latents, then decodes the corresponding action for max Q latent, and sends it to ARCLE. ARCLE executes the action and updates the state information accordingly. This process alternates between ARCLE and the inference network, continuing the inference until the episode ends.}
  \label{fig:total_flow}
\end{figure}

After training the agent using LDCQ on the SOLAR dataset, we conducted an evaluation of its performance. To evaluate our experiment, we synthesized an evaluation SOLAR set with 100 test examples, each paired with three synthesized demonstration examples. The evaluation SOLAR set was synthesized by the SOLAR-Generator using the same tasks but with a random seed different from the one used for the training set. To measure the effectiveness of decision-making using the Q-function, two accuracy metrics are measured: 1) Whether the agent reaches the answer state, and 2) Whether it predicts the \texttt{Submit} operation at the answer state to receive a reward.

The evaluation process is carried out through ARCLE, which manages the problem and its corresponding solution from SOLAR. ARCLE handles state transitions, performs actions, and verifies whether the submitted solution is correct. As depicted in Figure~\ref{fig:total_flow}, ARCLE interacts with the LDCQ inference network by alternating the exchange of $\bm{s}_t$ and $\bm{a}_t$, facilitating the decision-making process toward reaching the correct answer state. The latent $\bm{z}_{t}$ represents a segment trajectory spanning from timestep $t$ to $t+H-1$, and is trained to accurately decode actions for any state within this segment trajectory. 

In the original LDCQ methodology, inference is performed by executing several horizons using a single latent, followed by predicting the next latent. However, in the task used for this research, which has a gold standard trajectory consisting of five steps, it is possible to complete the task with just one latent sampling from the initial state. While reaching the correct answer in this manner is not inherently problematic, one of the primary goals of this research is to analyze whether the agent learns the knowledge prior to how actions work across various states. Thus, instead of focusing solely on solving the problem in as few steps as possible, only one action is conducted per latent. With this, the results demonstrate that the agent can make far-sighted decisions to reach the answer not just from the beginning to the end, but also through intermediate steps.

\subsection{Results}
To demonstrate the strengths of the diffusion-based offline RL method guided by Q-function, we compare three approaches:

\begin{itemize}
\item \textbf{VAE prior (VAE)}: This method uses a latent sampled from the VAE state prior \(p_{\omega}(\bm{z}_{t}|\bm{s}_t)\). The VAE state prior is trained in $\beta$-VAE training stage by calculating the KL divergence between \(p_{\omega}(\bm{z}_{t}|\bm{s}_t)\) and the posterior \(q_{\phi}(\bm{z}_{t}|\bm{\tau}_{t})\), aligning the latent distribution with the trajectory starting from state \(\bm{s}_t\).

\vspace{0.5em}
    
\item \textbf{Diffusion prior (DDPM)}: This method uses a latent sampled from the diffusion model \(p_\psi(\bm{z}_{t}|\bm{s}_{t})\) through the DDPM method~\citep{ho2020denoising}. The sampled latents closely resemble the training data, with added variance during the denoising process. This method is similar to behavior cloning in that it operates without guidance from rewards or value functions.
    
\vspace{0.5em}
    
\item \textbf{Max Q latent (LDCQ)}: This method selects a latent with the highest Q-value from those sampled by the diffusion model, $ \argmax_{\bm{z}\sim p_\psi(\bm{z}_{t}|\bm{s}_{t})} Q(\bm{s}_{t}, \bm{z})$, to make a decision at $\bm{s}_{t}$.
\end{itemize}

\begin{figure}[htb!]
  \hspace{-0.04\textwidth}
  \begin{subfigure}{0.44\textwidth}
    \centering
    \includegraphics[height=4cm] {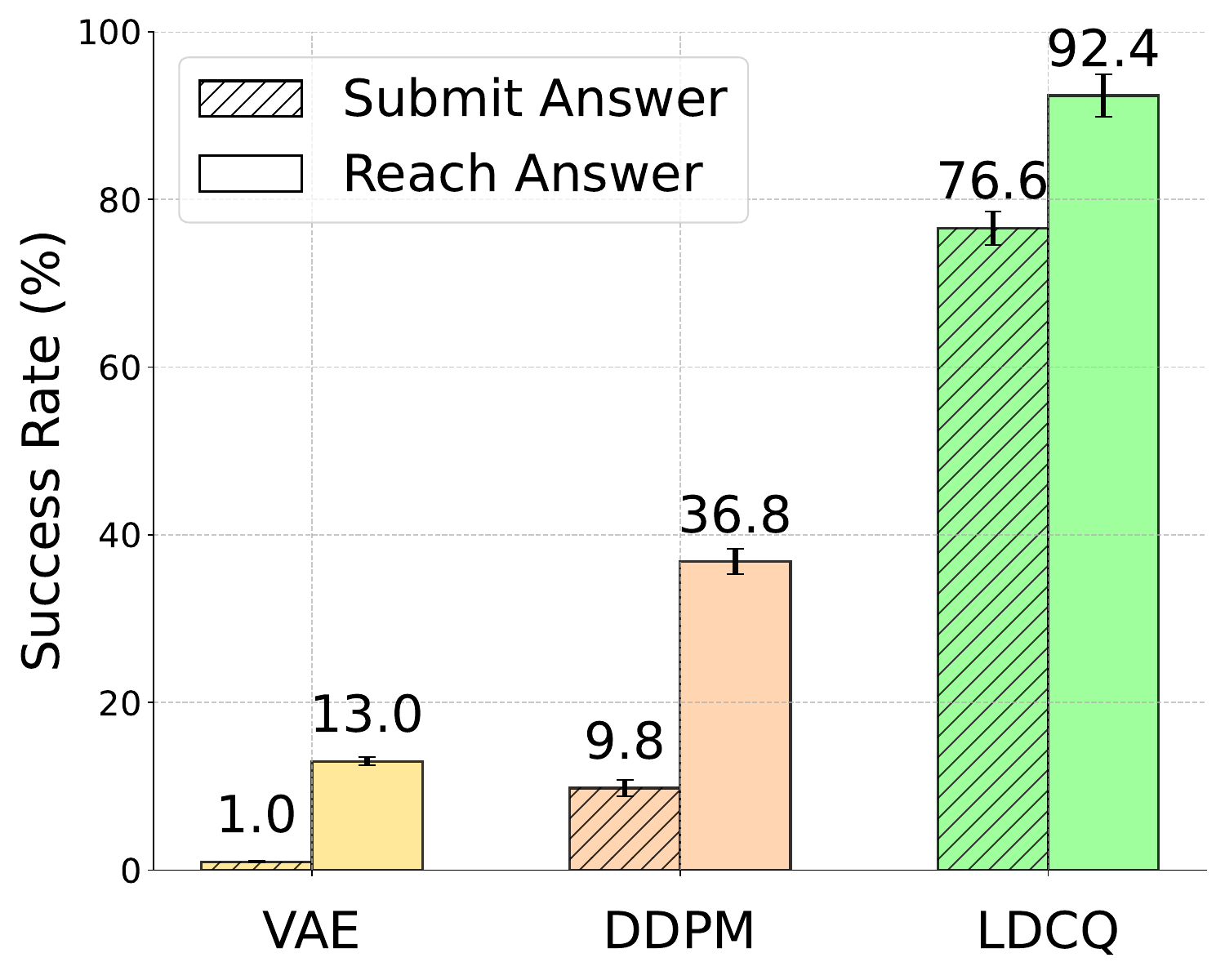}
    \caption{Test accuracy for three methods}
    \label{fig:compare_3}
  \end{subfigure}
  \hspace{-0.03\textwidth}
  \begin{subfigure}{0.53\textwidth}
    \centering
    \begin{subfigure}{\textwidth}
        \centering
        \includegraphics[height=1.75cm, keepaspectratio]{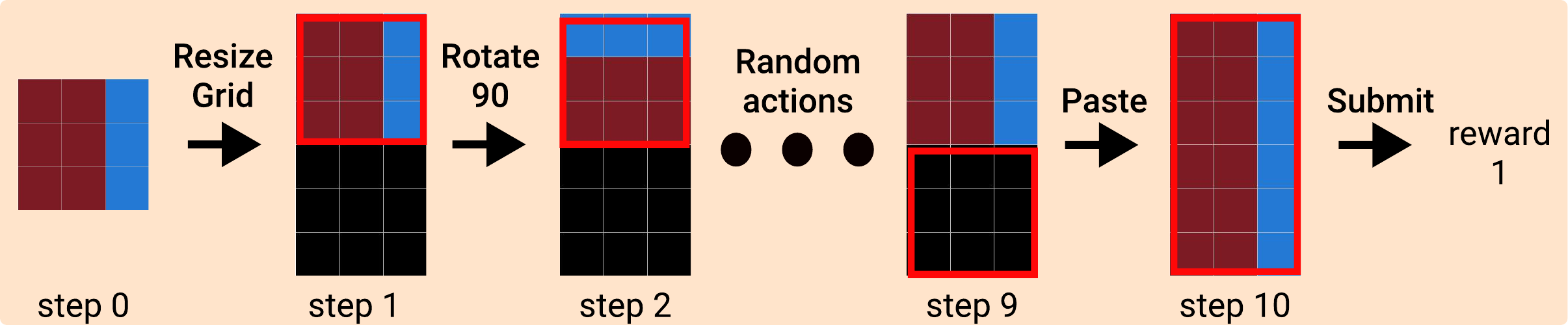}
        \caption{Inference example with DDPM method}
        \label{fig:diffusion_prior}
    \end{subfigure}
    \vfill
    \begin{subfigure}{\textwidth}
        \centering
        \includegraphics[height=1.75cm, keepaspectratio]{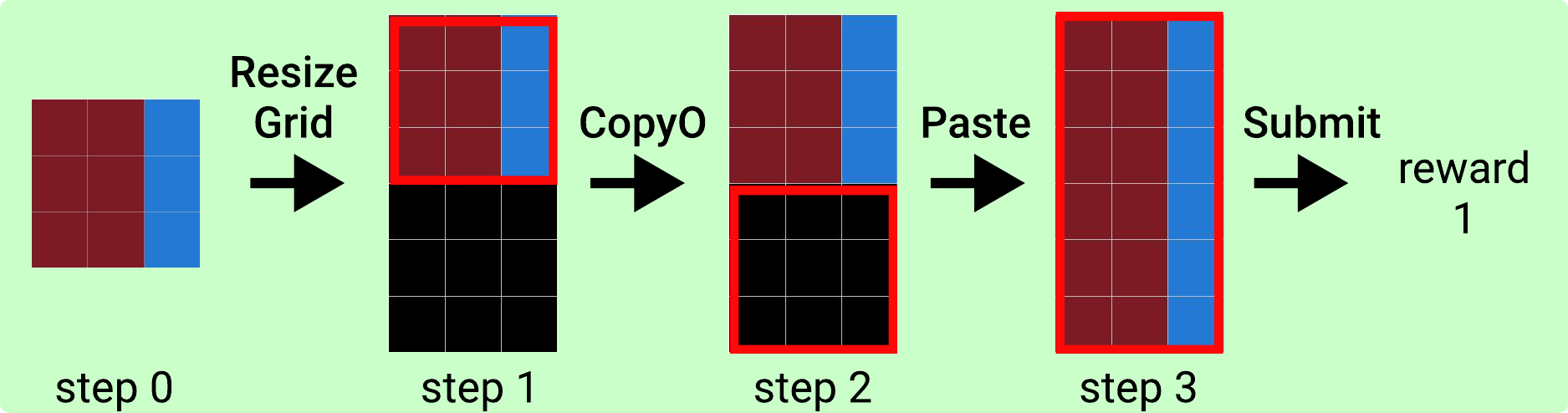}
        \caption{Inference example with LDCQ method}
        \label{fig:q}
    \end{subfigure}
  \end{subfigure}
  \caption{(a) The evaluation results for 100 test examples. LDCQ shows significantly improved performance compared to the other two methods, successfully reaching the correct answer state and executing the \texttt{Submit} operation at the answer state. The error bars represent the 96\% confidence interval. (b) With the latent sampled with DDPM, the agent sometimes reaches the correct answer after performing various actions. This occurred rarely during evaluation, and even when it did, it did not appear in subsequent evaluations. (c) When using LDCQ, it often shows the case that skips unnecessary actions. The inference example with the VAE prior method is omitted because it rarely solves the problem.}
  \label{fig:result}
\end{figure}

The evaluation of each approach was conducted five times for the evaluation SOLAR set. The results, summarized in Figure~\ref{fig:compare_3}, show the success rates for: 1) Whether the agent reaches the correct answer state and 2) Whether the agent executes \texttt{Submit} operation in the answer state. When using the VAE prior, the agent reaches the correct answer state in only about 10\% of test episodes and submits the answer in just 1\%. With latents sampled using DDPM, about 10\% of the answers are correctly submitted, while the agent reaches the answer state approximately 37\% of the time. When using LDCQ, the agent reaches the answer state in over 90\% of cases and successfully submits the correct answer in about 77\% of test episodes. These results demonstrate that the Q-function enhances the agent’s ability to both reach the correct answer and recognize when it has arrived at the answer state.

Figure~\ref{fig:diffusion_prior} and Figure~\ref{fig:q} highlight the different solving strategies exhibited by the Q-function. When using the latent sampled with DDPM, the agent performs diverse actions, occasionally reaching the goal by chance. In contrast, with the Q-function, the agent consistently reaches the correct answer in every evaluation. In scenarios where the input grid is vertically symmetrical, the agent even skips unnecessary operation \texttt{FlipV} and proceeds directly to \texttt{Submit}. Notably, the training dataset does not include any trajectories where the \texttt{FlipV} operation is skipped, even for symmetrical grids. With the Q-function, the model recognizes that applying \texttt{FlipV} does not alter the state. Consequently, the Q-value for submitting at that state increases, prompting the agent to choose the \texttt{Submit} operation. This demonstrates the reasoning ability of the agent trained with LDCQ in solving ARC tasks, as recognizing when the correct answer state has been reached is crucial.

\section{Limitations \& Discussions}

% 일단 앞 부분 분량이 줄지 않는다는 가정 하에, 8장을 맞춰보기 위하여 요약본을 붙여넣어 두었습니다. 다른 부분을 완성한 후에 분량을 최종 조절합시다.

In our experiment, the LDCQ method showed significant improvement in reaching the goal. However, in approximately 16\% of cases, the agent reached the correct state but proceeded with another action instead of submitting the solution, even with the assistance of the Q-function. This issue arises because the Q-function, while enhancing decision-making, sometimes assigns higher values to actions other than submission, causing the agent to bypass the goal state. This suggests that the Q-function is not perfectly aligned with the final objective in ARC. Notably, in ARC tasks, even when solving different test examples within the same task where the same rule is applied, the actual action sequence can vary depending on factors like grid size or the arrangement of elements in the input grid. The current Q-values are calculated based on the absolute state values, which occasionally leads to misjudgments when submitting the correct solution. Therefore, improving the agent’s ability to accurately determine when to submit the correct answer is necessary for future research.

While the LDCQ approach performs well in a simple ARC task setting, more complex tasks and multi-task environments present additional challenges. Unlike single-task scenarios, where the agent follows a fixed strategy toward a predefined answer, multi-task settings demand flexibility to adapt to changing goals or new possibilities during task execution. We expect that addressing these challenges could involve integrating task classifiers for Q-learning. Additionally, incorporating modules so that the agent can revise its strategy during task execution---adjusting based on evolving states or objectives rather than rigidly following the initial strategy---may enhance its adaptability.

In traditional supervised RL approaches, such as those described by \citet{ghugare2024closing}, stitching typically occurs only when the goal remains consistent across tasks. To address this limitation, we employed temporal data augmentation, which involves starting from an intermediate state near the goal and setting a new target. In SOLAR, this could be extended by using non-optimal paths as goals in non-optimal trajectories. However, in ARC, where goals are determined by demonstration pairs, augmenting all goals is impractical. More careful strategies are needed to enable stitching for entirely new goals not previously encountered. If methodologies are developed that can combine existing actions toward different goals, we expect that SOLAR will facilitate these combinations.

Going forward, refining how the Q-function evaluates states and actions will be crucial. To improve performance, especially in multi-task environments, incorporating mechanisms that not only assess the state and action in relation to the goal but also guide the agent toward the most effective path to achieve the ultimate objective will be beneficial. Recognizing the task's context and how close states are to the correct solution is essential for ensuring that the Q-function helps navigate toward the goal efficiently.

\section{Conclusion}
\label{Sec:Conlcusion}

This research demonstrates the potential of offline reinforcement learning (RL), particularly the Latent Diffusion-Constrained Q-learning (LDCQ) method, for efficiently sequencing and organizing actions to solve tasks in grid-based environments like the Abstraction and Reasoning Corpus (ARC). To our knowledge, this work is the first to tackle ARC using a diffusion-based offline RL model within a properly designed environment, guiding agents step-by-step toward correct solutions without generating the full ARC grid at once. Through training on SOLAR, we successfully applied and evaluated offline RL methods, showing that agents can learn to find paths to the correct answer state and recognize when they've reached it. This suggests that RL with a well-designed environment is promising for abductive reasoning problems, potentially reducing data dependency compared to traditional methods. As tasks become more complex, especially in multi-task settings, refining the Q-function to address unique reward structures is crucial, with multi-task environments requiring task-specific adaptations to account for varying states and rewards. Integrating modules like task classifiers or object detectors could enhance the agent's ability to dynamically adjust its strategy, promoting more flexible decision-making. This research opens new avenues for program synthesis in analogical reasoning tasks with RL environments, potentially integrating with analogy findings techniques (hypothesis search with LLMs) to handle a wider range of ARC tasks.

\newpage
% 최종본 제출할 때 넣으면 됩니다.
% \subsubsection*{Acknowledgements}
% This work was supported by the IITP (RS-2023-00216011, No. 2019-0-01842), the National Research Foundation (RS-2023-00240062), and GIST (AI-based Research Scientist Project) funded by the Ministry of Science and ICT, Korea. 

\bibliography{reference}
\bibliographystyle{iclr2025_conference}

% Checklist 채워야 됩니다
\newpage

\appendix

\section {Training Details}
\label{Appendix:Training Details}

\subsection{Latent Diffusion Constrained Q-Learning (LDCQ)}
\label{Appendix:LDCQ}

\paragraph{Training Latent Encoder and Policy Decoder} 

The first stage in training with LDCQ is to train a $\beta$-VAE that learns latent representations. In this stage, the $\beta$-VAE learns how actions are executed over multiple steps to change the state. With $\mathit{H}$-horizon latents, it becomes easier to capture longer-term changes in the state. We use SOLAR as the training dataset $\mathcal{D}$, which contains $\mathit{H}$-length segmented trajectories $\bm{\tau}_{_t}$. Each $\bm{\tau}_{t}$ consists of state sequences $\bm{s}_{t:t+H}=[\bm{s}_t, \bm{s}_{t+1}, ..., \bm{s}_{t+H-1}]$ and action sequences $\bm{a}_{t:t+H}=[\bm{a}_t, \bm{a}_{t+1}, ..., \bm{a}_{t+H-1}]$, along with additional information such as demonstration examples. As shown in Figure~\ref{fig:ldcq-a}, during the $\beta$-VAE training stage, the encoder $q_\phi$ is trained to encode $\bm{\tau}_{t}$ into the latent representation $\bm{z}_{t}$, and the low-level policy decoder $\pi_\theta$ is trained to decode actions based on the given state and latent. For example, given the latent $\bm{z}_{t}$ and a state from the segment trajectory, $\bm{s}_{t+h}$ where $ h \in [0,H)$, the policy decoder decodes the action $\bm{a}_{t+h}$ for $\bm{s}_{t+h}$. The $\beta$-VAE is trained by maximizing the evidence lower bound (ELBO), minimizing the loss in Eq.~\ref{eq:VAE_loss}. The loss consists of the reconstruction loss from the low-level policy decoder and the KL divergence between the approximate posterior $q_{\phi}(\bm{z}_{t}|\bm{\tau}_{t})$ and the prior $p_{\omega}(\bm{z}_{t}|\bm{s}_t)$.

\begin{equation}
    \resizebox{0.94\linewidth}{!}{$
    \mathcal{L}_{\mathrm{VAE}}(\theta,\phi, \omega) = -\mathbb{E}_{\bm{\tau}_{t}\sim \mathcal{D}} \left[ \mathbb{E}_{q_{\phi}(\bm{z}_{t}|\bm{\tau}_{t})} \left[ \displaystyle\sum_{l=t}^{t+H-1} \log \pi_{\theta}(\bm{a}_l|\bm{s}_l, \bm{z}_{t}) \right]-\beta D_{KL}(q_{\phi}(\bm{z}_{t}|\bm{\tau}_{t}) \parallel p_{\omega}(\bm{z}_{t}|\bm{s}_t) ) \right]
    \label{eq:VAE_loss}
    $}
\end{equation}

\paragraph{Training Latent Diffusion Model} 

In the second stage, latent diffusion model is trained to generate latents based on the latent representations encoded by the $\beta$-VAE. The training data consists of $(\bm{s}_t, \bm{z}_{t})$ pairs, which are used to train a conditional latent diffusion model $p_\psi(\bm{z}_{t}|\bm{s}_{t})$ by learning the denoising function $\mu_\psi(\bm{z}_{t}^{j}, \bm{s}_{t}, j)$, where $j\in [0,T]$ is diffusion timestep. This allows the model to capture the distribution of trajectory latents conditioned on $\bm{s}_t$. $q(\bm{z}_{t}^{j} | \bm{z}_{t}^{0})$ denotes the forward Gaussian diffusion process that noising the original data. Following previous research~\citep{ramesh2022hierarchical,venkatraman2024reasoning}, we predict the original latent rather than the noise, balancing the loss across diffusion timesteps using the Min-SNR-$\gamma$ strategy~\citep{hang2023efficient}. The loss function used to train the diffusion model is shown in Eq.~\ref{eq:latent_diffusion_loss}. Here, $\bm{z}_{t}^{j},\,j \in [0,T]$ represents noised latent on $j$-th diffusion time step, when $j=0$ then $\bm{z}_{t}^{0}=\bm{z}_t$ and $\bm{z}_{t}^{T}$ is Gaussian noise. 

\begin{equation}
    \mathcal{L}(\psi) = \mathbb{E}_{j \sim [1,T], \bm{\tau}_H \sim \mathcal{D}, \bm{z}_{t} \sim q_\phi (\bm{z}_{t} | \bm{\tau}_{t}), \bm{z}_{t}^{j} \sim q(\bm{z}_{t}^{j} | \bm{z}_{t}^{0})} \bigl[ \min\{\mathrm{SNR}(j),\gamma\}\|\bm{z}_{t}^{0} - \mu_\psi(\bm{z}_{t}^{j}, \bm{s}_{t}, j)\|^2 \bigr]
    \label{eq:latent_diffusion_loss}
\end{equation}

\paragraph{Training Q-Network} 

Finally, the latent vectors sampled by the latent diffusion model are used for Q-learning. For sampling latents, we use the DDPM method~\citep{ho2020denoising}. The trained diffusion model samples latents by denoising random noise using the starting state information $\bm{s}_t$. We use the data consisting of ($\bm{s}_{t}, \bm{z}_{t}, r_{t:t+H}, \bm{s}_{t+H}$) for training Q-Network, where $r_{t:t+H}=\sum_{l=t}^{t+H-1}\gamma^l r_{l}$ deontes the discounted sum of rewards. Here, DDPM sampling is used to sample $\bm{z}_{t+H}$ for $\bm{s}_{t+H}$. For Q-learning, we use Clipped Double Q-learning~\citep{fujimoto2018doubleq} as shown in Eq.~\ref{eq:q_learning} with Prioritized Experience Replay buffer~\citep{schaul2015prioritized} to improve learning stability and mitigate overestimation. The trained Q-network $Q(\bm{s}_t, \bm{z}_{t})$ evaluates the expected return of performing various $H$-length actions, with $\bm{z}_{t}$ sampled via DDPM based on $\bm{s}_t$. This allows the network to efficiently calculate the value of actions over $H$-steps to estimate future returns.

\begin{equation}
    Q(\bm{s}_t, \bm{z}_{t}) \leftarrow \left( r_{t:t+H} + \gamma^H Q(\bm{s}_{t+H}, \argmax_{\bm{z}\sim p_\psi(\bm{z}_{t+H}|\bm{s}_{t+H})} Q(\bm{s}_{t+H}, \bm{z}) ) \right)
    \label{eq:q_learning}
\end{equation}

% More details for training with LDCQ are provided in Appendix~\ref{Appendix:Training Details}.

\newpage
\subsection{Hyperparameters}

We used a horizon length of 5 for encoding skill latents, meaning the model plans and evaluates actions over a five-step lookahead.

% In ARCLE, once the grid is submitted for evaluation, the episode ends when the state matches the correct answer~\cite{lee2024arcle}. Without the inclusion of 35 (\texttt{None}) operations, the model would only generate latents that continue performing other actions, delaying \texttt{Submit} even when in the correct state. To help the model efficiently recognize when the current state is correct and prompt it to execute the \texttt{Submit} operation, we added \texttt{None} operations after the \texttt{Submit}. This ensures that all segment trajectories have a horizon of 5, and for segments that included \texttt{Submit}, the remaining horizon was padded with \texttt{None} operations until it was filled.

We trained the diffusion model with 500 diffusion steps. If the number of diffusion steps is too small, it can lead to high variance in the sampling process, potentially causing errors during the decoding of operations or selections in ARCLE. To minimize these errors, we set the number of diffusion steps to 500, ensuring more accurate operation and selection decoding from the sampled latents.

We set the discount factor to 0.5 to ensure the model appropriately balances immediate and future rewards. Since the total steps required to reach the correct answer in ARCLE are usually fewer than 20, a high discount factor could cause the agent to struggle in distinguishing between submitting at the correct state and continuing with additional steps, which could lead to episode failure.

The hyperparameters that we used for training three stages of LDCQ are shown in Tables~\ref{tab:ARC_VAE_hyperparameter}, \ref{tab:ARC_Diffusion_hyperparameter} and \ref{tab:ARC_Q_hyperparameter}.

% VAE 하이퍼파라미터
\begin{table}[H]
\centering
\caption{Hyperparameters for training $\beta$-VAE }
\label{tab:ARC_VAE_hyperparameter}
\begin{tabular}{@{}lc@{}}
\toprule
\textbf{Parameter}                  & \textbf{Value} \\ 
\midrule
Learning rate                       & 5e-5 \\ 
Batch size                          & 128 \\ 
Epochs                              & 400 \\ 
Horizon ($H$)                       & 5 \\ 
Latent dimension ($\text{z}$)       & 256 \\ 
KL loss ratio ($\beta$)             & 0.1 \\ 
Hidden layer dimension              & 512 \\ 
\bottomrule
\end{tabular}
\end{table}

% Diffusion 하이퍼파라미터
\begin{table}[H]
\centering
\caption{Hyperparameters for training latent diffusion model}
\label{tab:ARC_Diffusion_hyperparameter}
\begin{tabular}{@{}lc@{}}
\toprule
\textbf{Parameter}                          & \textbf{Value} \\ 
\midrule
Learning rate                               & 1e-4 \\ 
Batch size                                  & 32 \\ 
Epochs                                      & 400 \\ 
Diffusion steps ($T$)                       & 500 \\ 
Drop probability                            & 0.1 \\ 
Variance schedule                           & linear \\ 
Sampling algorithm                          & DDPM \\ 
$\gamma$ (For Min-SNR-$\gamma$ weighing)     & 5 \\ 
\bottomrule
\end{tabular}
\end{table}

% Q 하이퍼파라미터
\begin{table}[H]
\centering
\caption{Hyperparameters for training DQN}
\label{tab:ARC_Q_hyperparameter}
\resizebox{\linewidth}{!}{%
\begin{tabular}{@{}lc@{}}
\toprule
\textbf{Parameter}                          & \textbf{Value} \\ 
\midrule
Learning rate                               & 5e-4 \\ 
Batch size                                  & 128 \\ 
Discount factor ($\gamma$)                  & 0.5 \\ 
Target net update rate ($\rho$)             & 0.995 \\ 
PER buffer $\alpha$                         & 0.7 \\ 
PER buffer $\beta$                          & Linearly increased from 0.3 to 1, Grows by 0.03 every 2000 steps \\ 
Diffusion samples for batch $\argmax$          & 100 \\ 
\bottomrule
\end{tabular}}
\end{table}

\subsection{Hardware}
\label{Appendix:Hardware}

We used an NVIDIA A100-SXM4-40GB GPU to train the model. Training the $\beta$-VAE took about 7 hours, while training the diffusion model and Q-network each took around 6 to 10 hours.

\section{Details of SOLAR-Generator}
\label{Appendix:SOLAR}

\subsection{Operations in SOLAR}
\label{Appendix:operations in SOLAR}

\begin{figure}[h!]
    \centering
    \includegraphics[width=0.8\linewidth]{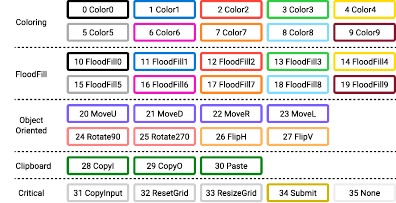}
    \caption{All operations compatible with SOLAR, 0\textendash34 operations follow ARCLE, and only in SOLAR, 35 (\texttt{None}) is for terminated episode. It means the episode is ended after \texttt{Submit}. }
    \label{fig:solar-actions}
\end{figure}

The operations from 0 to 34 are identical to those used in ARCLE~\citep{lee2024arcle}. Since \texttt{Submit} is an operation that receives a reward, it should only be used when the state is considered correct and not excessively. Due to LDCQ's fixed horizon, and to ensure that the agent only uses \texttt{Submit} when the state is definitively correct, we added a \texttt{None} operation that fills all subsequent states after \texttt{Submit} with the 11th color (10), which does not exist in the original ARC (0–9). In other words, during training, the \texttt{None} action emphasizes that the episode ends after \texttt{Submit}.

% \subsection{Detailed Procedure for Generating SOLAR}
\label{Appendix:SOLAR-Generator}
\subsection{Grid Maker}
For generating SOLAR, we create a generator that can synthesize a large amount of data for a given rule. Grid Maker is a hard-coded program specific to each task. \text{Grid Maker} contains the rules for synthesizing demonstration examples and test examples, and the synthesized solution action path consists of operations and selections. In \text{Grid Maker}, data is formatted to be compatible with ARCLE. The Grid Maker constructs analogies with the same problem semantics but with various attributes such as the shape, color, size, and position of objects. SOLAR-Generator can generate intermediate trajectories by interacting with ARCLE. The algorithm of the SOLAR-Generator is designed to augment specific tasks using the \text{Grid Maker}, which can primarily be divided into three parts.

Grid Maker was built as a data loader, which is used in ARCLE. In the original ARCLE environment, there was no need to load operations and selections—only the grid was loaded since the problem alone was sufficient. To change this structure, the entire environment would need to be recreated. Instead, operations and selections are now loaded from the data loader’s description, allowing us to retain the original environment. Therefore, the process of creating input-output examples and generating action sequences works within a single file.

\paragraph{Specifying Common Parts} Each task in the ARC dataset usually contains 3 demonstration examples, with common elements observed across these pairs. In the common parts, attributes such as color, the type of task, and the presence of objects are predetermined using random values before pair generation.

\paragraph{Synthesizing Examples} In the example synthesis phase, the input of the original task is augmented in a way that ensures diversity while preserving the integrity of the problem-solving method. A random input grid is generated under conditions that satisfy the analogy required by the task. A solution grid is created using a hard-coded algorithm. For tasks involving pattern-based problems, as experimented in the paper, selections are made to fit the grid size, and various operations are executed either randomly or in a predetermined order. For object-based problems, the solution grid is generated by an algorithm that finds the necessary objects in the input grid and processes them according to the task requirements.

\paragraph{Converting to ARCLE Trajectories} This stage involves the creation of an ARCLE-based trajectory that meticulously adheres to the problem-solving schema of the synthesized examples. The entire process is carried out through a hard-coded algorithm. During the example synthesis process, the locations of objects may already be known, or they can be identified using a search algorithm. The information obtained is then used to make the appropriate selections, and the trajectory is converted into an ARCLE trajectory through an algorithm that leads to the correct solution. 

If all steps are properly coded, it is possible to generate the operations and selections that lead to the correct solution for any random input grid. These are then fed into ARCLE to obtain intermediate states, rewards, and other information, and to verify whether the correct result is reached. Once steps 1) to 3) are correctly implemented, SOLAR-Generator can continuously and automatically generate as much data for the given task as the user desires, using the Grid Maker.

\subsection{Example of Data Synthesis in Grid Maker and the Generation of SOLAR}
\begin{figure}[H]
    \centering
    \includegraphics[width=0.58\linewidth]{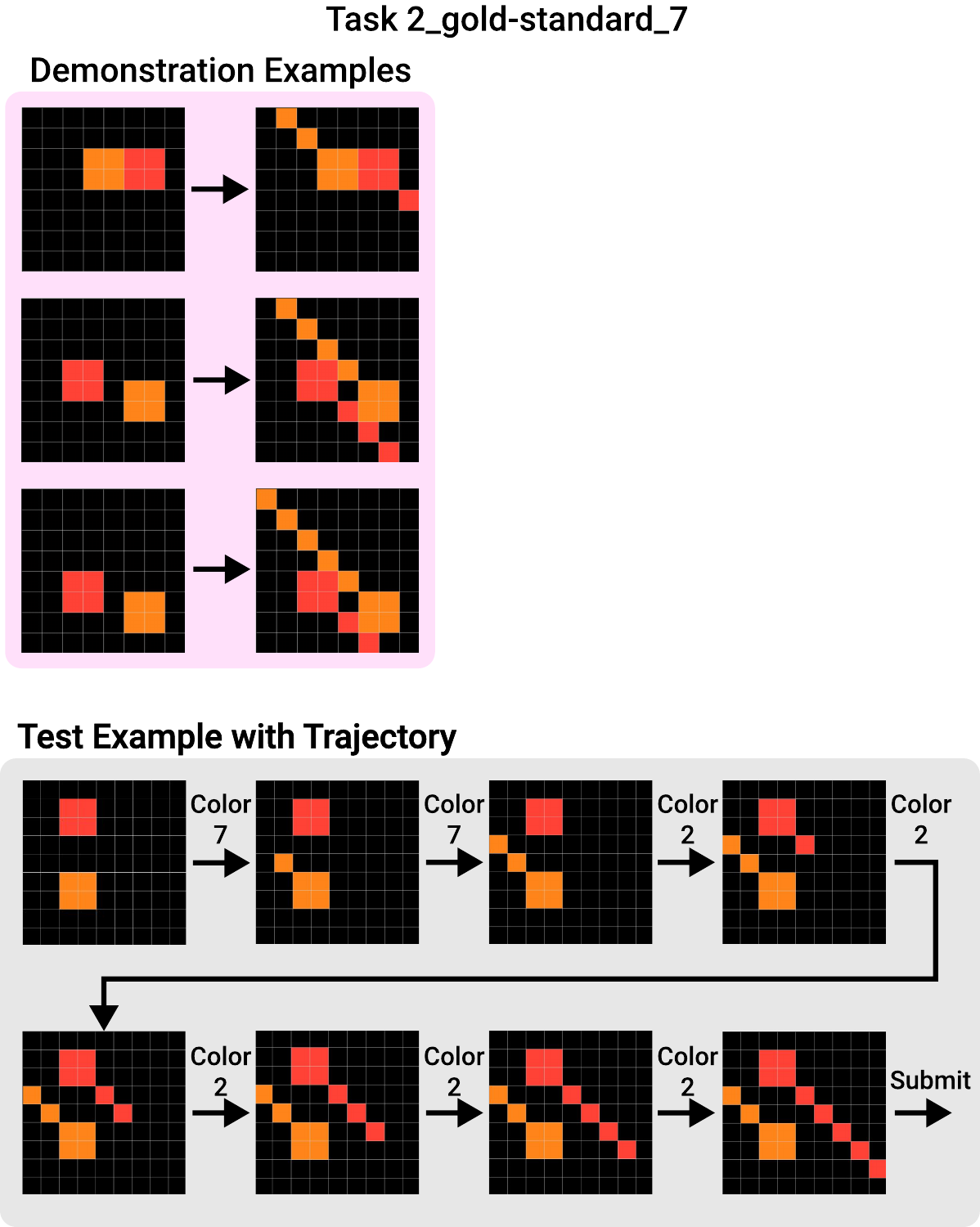}
    \caption{A gold standard trajectory for Task 2 in Figure~\ref{ARC-Example}. SOLAR contains its trajecotry ID, demonstration examples, and a test example with trajectory.}
    \label{fig:task2}
\end{figure}

SOLAR-Generator can synthesize SOLAR for object-based tasks. Figure~\ref{fig:task2} shows a variant of Task 2 from Figure~\ref{ARC-Example}. Grid Maker generates random input grids with some variances firtst. In this variant, each episode randomly selects two colors for the boxes. Each inputs can have different grid sizes, and rules are established for objects of each color within the episode. Then it generates the answer output grids for the input grids through algorithm. The solution algorithm in Grid Maker proceeds as follows: 1) Find the top-left corner of the orange square and repeat the coloring process to draw a diagonal line to the grid's edge. 2) Find the bottom-right corner of the red square and repeatedly color diagonally until the end of the grid is reached. With these algorithms, Grid Maker can synthesize as many examples and SOLAR trajectories as the user desires.

\newpage
\subsection{Algorithm of SOLAR-Generator}
With the synthesized data through the Grid Maker module, the SOLAR-Generator checks the sanity of the synthesized trajectory, and then saves the data. The whole algorithm for SOLAR-Generator is described in Algorithm~\ref{alg:SOLAR-Generator}.

\begin{algorithm}[H]
\caption{SOLAR-Generator}
\label{alg:SOLAR-Generator}

\textnormal{Input:} task set \textit{T}, maximum grid size (\textit{H,W}), number of samples \textit{N}, number of examples \textit{E}

\medskip

\For{ \textit{task} $\in$ \textit{T}}{

    % Load task-specific grid maker
    \# Load the synthesized data $\mathcal{D}_s$ from the Grid Maker for the \textit{task}

    $\mathcal{D}_s$ $\gets$ Grid Maker($\textit{task} ,\,(\textit{H},\,\textit{W}),\, \textit{N},\,\textit{E}$)
        
    \For{ \textit{data} $\in \, \mathcal{D}_s$ } { 
    
        % Extract initial state, operations, selections
        \# Extract the demonstration examples, test example, and actions for each episode
    
        \textit{trajectory\_ID, dem\_ex, input\_grid, output\_grid, operations, selections} $\gets$ \textit{data}
        \medskip

        Add \textit{trajectory\_ID,  dem\_ex, input\_grid, output\_grid} to episode $\tau_{\textit{data}}$ 
       
       \medskip
        % Set the initial state
        \# Set the initial state
        
        $\textit{current\_grid}_0 \gets$ \textit{input\_grid}
        
        $\textit{clip\_grid}_0 \gets None$
       
        $t \gets 0$

        $s_t \,\gets \, (\textit{input\_grid}, \textit{current\_grid}_0, \textit{clip\_grid}_0)$
        
        \medskip

        \For{$(opr_t, sel_t)$ $\in$ \textit{(operations, selections)}} {
            
           $a_t \gets (opr_t, sel_t) $
           
            \If{$a_t$ \textnormal{can be performed \textnormal{in}} $s_t$}{
                \# Update state and episode information using ARCLE
                
                $\textit{current\_grid}_{t+1}, \textit{clip\_grid}_{t+1}, \textit{reward}_t, \textit{terminated}_t \gets \texttt{ARCLE.step}(s_t, a_t)$

                Add $s_t, a_t,\textit{reward}_t, \textit{terminated}_t$ to $\tau_{\textit{data}}$
                
                $s_{t+1} \,\gets \, (\textit{input\_grid}, \textit{current\_grid}_{t+1}, \textit{clip\_grid}_{t+1})$
                
                $t \gets t + 1$
            }         
            \Else{
                    Save wrong data for debugging
                    
                    \textnormal{break}
            }
        }

       \If {``\textit{gold-standard}'' \textnormal{\textnormal{in}} \textit{trajectory\_ID} \textnormal{\textnormal{and}} \textit{current\_grid} $\ne$ \textit{output\_grid}}{
                Save wrong data for debugging
            }
                
        \Else{
                Save episode $\tau_{\textit{data}}$
            }
    }
}
\end{algorithm}

\newpage
\subsection{Other SOLAR Examples}
\label{Appendix:SOAR-examples}

\begin{figure}[htb]
    \centering
    \begin{subfigure}[b]{1.0\linewidth}
        \centering
        \includegraphics[width=0.51\linewidth]{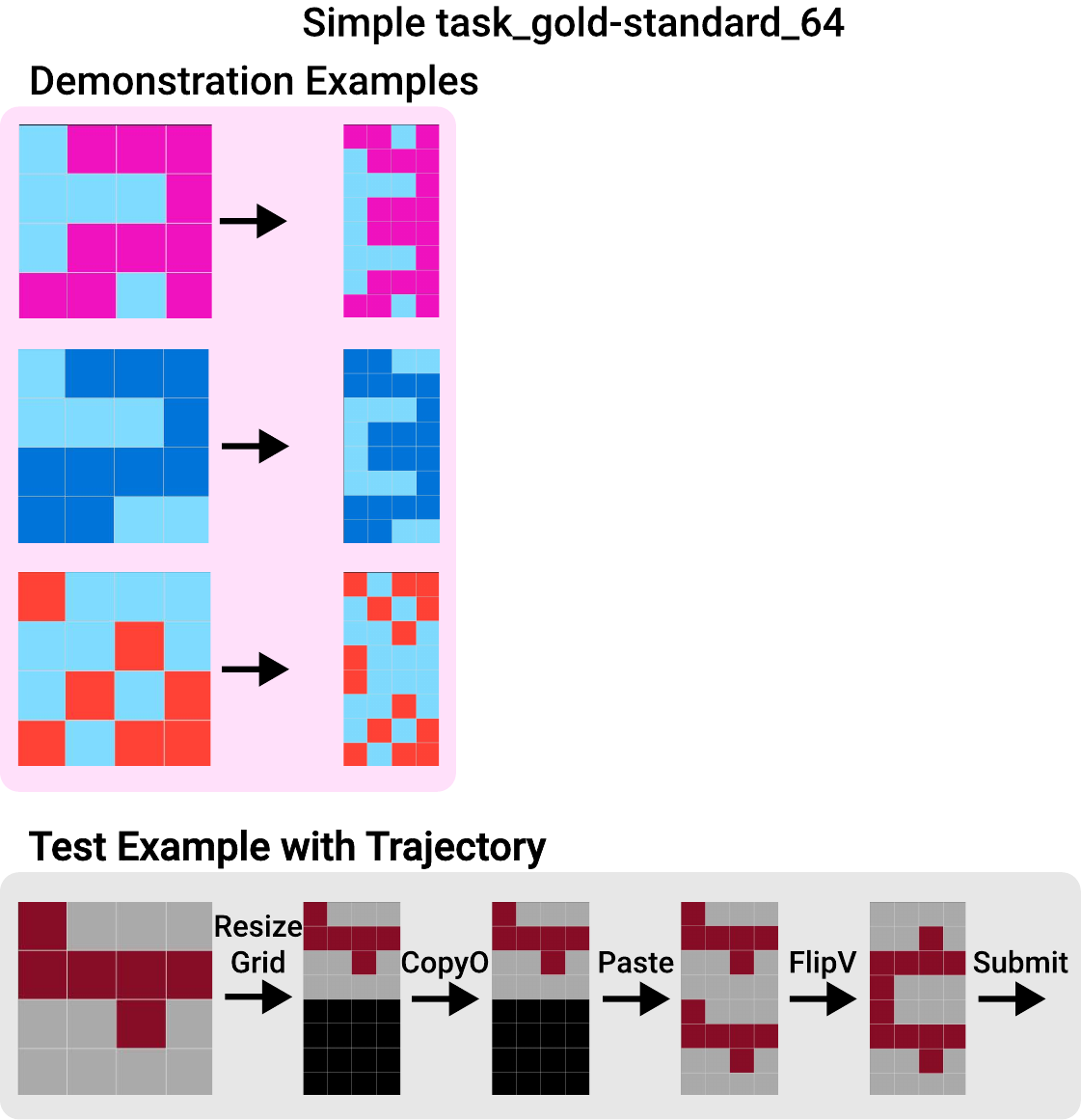}
        \caption{Gold standard episode}
        \label{fig:simple_task_a}
    \end{subfigure}
    \hspace{0.04cm}
    \begin{subfigure}[b]{1.0\linewidth}
        \centering
        \includegraphics[width=0.61\linewidth]{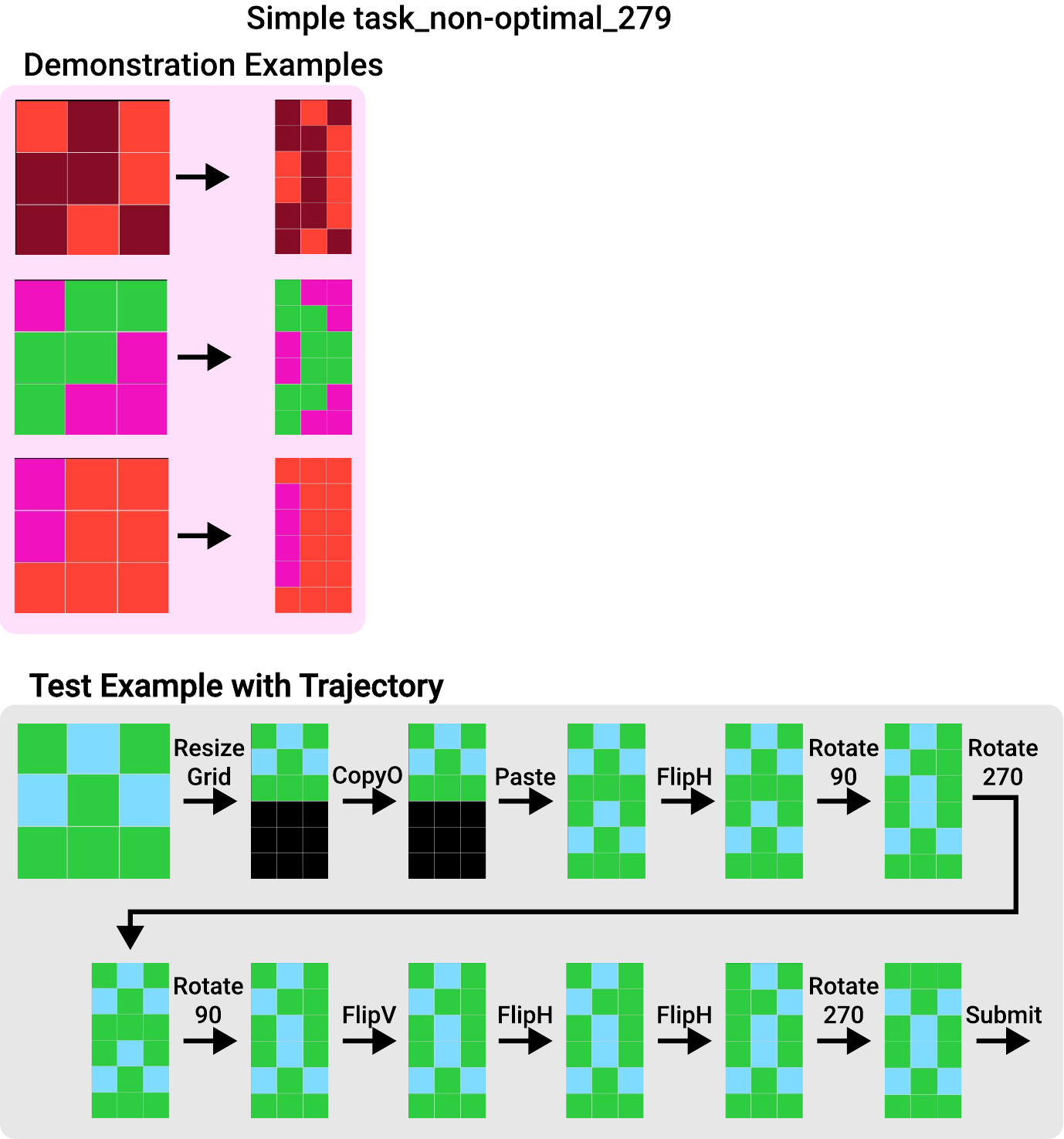}
        \caption{Non-optimal episode}
        \label{fig:simple_task_b}
    \end{subfigure}
    \caption{SOLAR episode examples of the task used in our experiment. Each episode contains three demonstration examples and a test example with a trajectory. (a) An example of a gold standard episode that ideally reaches the answer. (b) An example of a non-optimal episode that is not ideal, but still reaches the answer state.}
    \label{fig:simple_task}
\end{figure}

Figure~\ref{fig:simple_task} illustrates two examples of episodes from the tasks used in the experiment. Each episode includes three random demonstration examples and a trajectory for a test example. Figure~\ref{fig:simple_task_a} shows a gold standard trajectory, which represents the ideal sequence of actions to reach the correct answer state. Figure~\ref{fig:simple_task_b} shows a non-optimal trajectory that, while not a gold standard, also reaches the answer state. The clip grid, reward, and termination information are not displayed.

\newpage
\begin{figure}[H]
    \centering
    \includegraphics[width=\linewidth]{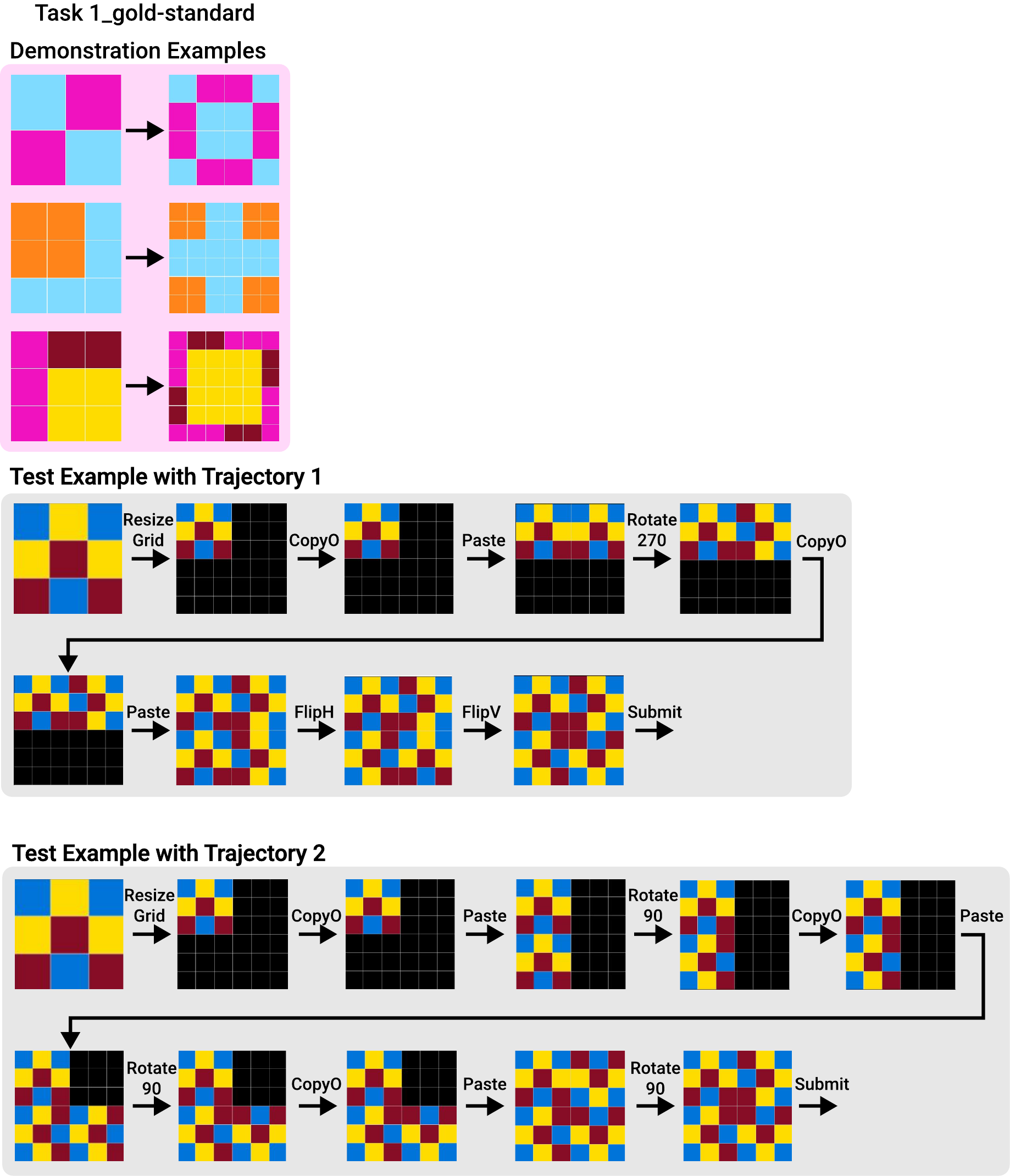}
    \caption{Two different gold standard trajectories for Task 1 in Figure~\ref{ARC-Example}, there might be multiple gold standard trajectories in the same task.}
\end{figure}

\newpage

\section{SOLAR-Generator vs RE-ARC: Comparing ARC Data Augmentation Approaches}

During the development of SOLAR, another augmentation scheme called RE-ARC~\citep{hodel2024rearc} was independently developed and released. While both aim to generate diverse examples for ARC tasks, they differ significantly in their design objectives, underlying architectures, and dataset structures.

\paragraph{Underlying Architectures}
SOLAR is built upon the ARCLE framework, designed for training reinforcement learning agents. It uses a limited set of actions based on the O2ARC web interface, which, despite their simplicity, are sufficient primitives to solve all ARC tasks. This design choice results in sequential trajectories directly applicable to reinforcement learning models. 
In contrast, RE-ARC is based on a more comprehensive Domain Specific Language (DSL) developed by Hodel, featuring 141 primitives. This expanded set of operations provides greater flexibility in expressing solutions, allowing for more complex transformations.

\paragraph{Data Generation Approach}
SOLAR generates sequential trajectories that mirror the step-by-step approach humans use when solving ARC tasks. This aligns well with typical reinforcement learning models that execute actions sequentially.
RE-ARC, leveraging its expansive DSL, generates solutions in the form of directed acyclic graphs (DAGs). This approach allows for more complex problem-solving strategies but may present challenges when applied to traditional reinforcement learning frameworks.

\paragraph{Dataset Structure and Utility}
SOLAR provides complete episodes with detailed trajectories, including all intermediate states. This feature is crucial for training agents with offline reinforcement learning methods, allowing models to learn from the entire problem-solving process. 
RE-ARC focuses on augmenting input-output pairs without explicitly providing intermediate steps. While valuable for increasing example diversity and testing generalization capabilities, it may require additional processing for direct application in a reinforcement learning context.

\paragraph{Flexibility and Potential for Integration}
SOLAR's design allows for easy generation of large episode datasets and is highly adaptable for various experimental setups in reinforcement learning research. The simplicity of the ARCLE action set makes it easier to modify and extend the system. 
RE-ARC's DAG-based approach, while not immediately compatible with sequential RL methods, opens up possibilities for more advanced RL frameworks capable of handling DAG-structured data.

\paragraph{Future Directions}
Future research could explore synergies between SOLAR and RE-ARC approaches, potentially leading to more powerful and flexible AI systems for solving ARC tasks. One promising direction would be adapting the LDCQ methodology to work with RE-ARC's DAG structures, which could involve developing new RL algorithms capable of processing DAG-structured data.
Another interesting avenue would be to investigate how SOLAR's sequential trajectories could inform or constrain the generation of more complex DAG-based solutions in RE-ARC. Such a hybrid approach could combine the simplicity and learnability of sequential actions with the expressiveness of DAG-based representations.
By integrating the strengths of both approaches - SOLAR's alignment with current RL techniques and RE-ARC's comprehensive problem representations - we could potentially unlock new capabilities in AI systems. This integration might lead to significant advancements in abstract reasoning and problem-solving, bridging the gap between the efficiency of reinforcement learning and the expressiveness of symbolic methods.

% \section{Difference from Another Augmentation Scheme}

% During the development of SOLAR, the RE-ARC~\citep{hodel2024rearc} was released. RE-ARC provides code to procedurally generate diverse examples for the 400 ARC tasks by reverse-engineering the underlying distribution of examples.

% RE-ARC and SOLAR are significantly different in their design objectives and dataset structure. The biggest difference between SOLAR and RE-ARC is that SOLAR includes episodes with intermediate steps, whereas RE-ARC focuses solely on augmenting input-output pairs. These intermediate steps in SOLAR are crucial for training agents with offline RL methods, as they provide detailed trajectories leading to the solution or not. Consequently, SOLAR offers data specifically designed for solving ARC tasks using offline RL methods. By modifying the Grid Maker, SOLAR can easily generate as many episode datasets as needed, even when adding variance to existing tasks, making it highly flexible for experimentation.

% \label{Appendix:Detialed Results}
% \input{}

\newpage

%제출시에 supplemenray 앞으로 보내야합니다!!!!
% \input{checklist}
\end{document}